\documentclass[numbers]{article}
\usepackage[table,xcdraw]{xcolor} 
\usepackage[final]{neurips_2025}

\usepackage[utf8]{inputenc} % allow utf-8 input
\usepackage[T1]{fontenc}    % use 8-bit T1 fonts
\usepackage{hyperref}       % hyperlinks
\usepackage{url}            % simple URL typesetting
\usepackage{booktabs}       % professional-quality tables
\usepackage{amsfonts}       % blackboard math symbols
\usepackage{nicefrac}       % compact symbols for 1/2, etc.
\usepackage{microtype}      % microtypography

\usepackage{graphicx}

\usepackage{amsmath}
\usepackage{amsthm}
\usepackage{caption}
\usepackage{multirow}
\usepackage{adjustbox}
\usepackage{amssymb}
\usepackage{mathtools}
\usepackage{algorithm}
\usepackage{algpseudocode}
\theoremstyle{plain}
\usepackage{wrapfig}
\usepackage{adjustbox}  
\usepackage{amssymb}
\usepackage{nicematrix}
\usepackage{booktabs}  
\newtheorem{theorem}{Theorem}[section]
\newtheorem{proposition}[theorem]{Proposition}

\theoremstyle{definition}

\theoremstyle{remark}

\usepackage{dblfloatfix} 
\usepackage{float}
\definecolor{deblue}{RGB}{11,132,147}
\definecolor{ocra}{RGB}{204, 119, 34}
\usepackage{enumitem}
\newcommand{\fcircle}[2][red,fill=red]{\tikz[baseline=-0.5ex]\draw[#1,radius=#2] (0,0.03) circle ;}
\usepackage{colortbl} % 第二步
\usepackage{nicematrix} 
\usepackage{tikz} 
\usepackage[textsize=tiny]{todonotes}

\title{NeuTSFlow: Modeling Continuous Functions Behind Time Series Forecasting}
\author{%
  \bf Huibo Xu\textsuperscript{1},
  Likang Wu\textsuperscript{2},
  Xianquan Wang\textsuperscript{1},
  Haoning Dang\textsuperscript{3}, 
  Chun-Wun Cheng\textsuperscript{4},\\
  \bf Angelica I Aviles-Rivero\textsuperscript{5},
  Qi Liu\textsuperscript{1}\thanks{Corresponding author: \texttt{qiliuql@ustc.edu.cn}}  \\
  \small{\textsuperscript{1}University of Science and Technology of China} \\
  \small{\textsuperscript{2}Tianjin University} \\
  \small{\textsuperscript{3}Xi’an Jiaotong University} \\
  \small{\textsuperscript{4}University of Cambridge} \\
  \small{\textsuperscript{5}YMSC, Tsinghua University} \\[3pt]
}

\begin{document}

\maketitle

\begin{abstract}
Time series forecasting is a fundamental task with broad applications, yet conventional methods often treat data as discrete sequences, overlooking their origin as noisy samples of continuous processes. Crucially, discrete noisy observations cannot uniquely determine a continuous function; instead, they correspond to a family of plausible functions. Mathematically, time series can be viewed as noisy observations of a continuous function family governed by a shared probability measure. Thus, the forecasting task can be framed as learning the transition from the historical function family to the future function family. This reframing introduces two key challenges: (1) How can we leverage discrete historical and future observations to learn the relationships between their underlying continuous functions? (2) How can we model the transition path in function space from the historical function family to the future function family? To address these challenges, we propose NeuTSFlow, a novel framework that leverages Neural Operators to facilitate flow matching for learning path of measure between historical and future function families. By parameterizing the velocity field of the flow in infinite-dimensional function spaces, NeuTSFlow moves beyond traditional methods that focus on dependencies at discrete points, directly modeling function-level features instead. Experiments on diverse forecasting tasks demonstrate NeuTSFlow's superior accuracy and robustness, validating the effectiveness of the function-family perspective.

\end{abstract}

\section{Introduction}

Time series forecasting is a fundamental task in time series analysis, with critical applications spanning healthcare~\cite{kaushik2020ai}, climate science~\cite{ray2021time}, and industrial systems~\cite{fatima2024review}. Current forecasting methods~\cite{liu2023itransformer,zeng2023transformers,chen2021autoformer} typically treat time series data as discrete sequences, focusing on learning mappings between historical and future observations~\cite{kerrigan2023functionalflowmatching,tong2024improvinggeneralizingflowbasedgenerative,lipman2023flowmatchinggenerativemodeling,albergo2023stochasticinterpolantsunifyingframework,albergo2023buildingnormalizingflowsstochastic,liu2022flowstraightfastlearning,pooladian2023multisampleflowmatchingstraightening}. This approach overlooks a key characteristic of real-world time series: they are discrete, noisy observations of underlying continuous processes, such as electricity load, wind speed, or temperature~\cite{sheoran2022wind}.
This naturally leads to the idea that if a model can learn the relationships between the functions underlying these discrete sequences, it would be able to capture the more fundamental characteristics of the underlying processes. Mathematically, however, discrete observations cannot uniquely determine an underlying continuous function. Even without noise, countless continuous functions can fit the same discrete points. When noise is present, this ambiguity increases, meaning a discrete time series represents not a single continuous function but a family of plausible functions consistent with the noisy observations.

Thus, time series prediction can be reframed as learning the transition from the historical function family to the future function family. This reframing introduces two key challenges: (1) \textbf{How can we model the transition path in function space from the historical function family to the future function family?} While the transition between function families must be learned based on the relationships between their corresponding continuous functions, only discrete observations are available. \textbf{(2) How can we leverage discrete historical and future observations to learn the relationships between their underlying continuous functions? }

\begin{figure}[t!]
\centering
\includegraphics[width=1\textwidth]{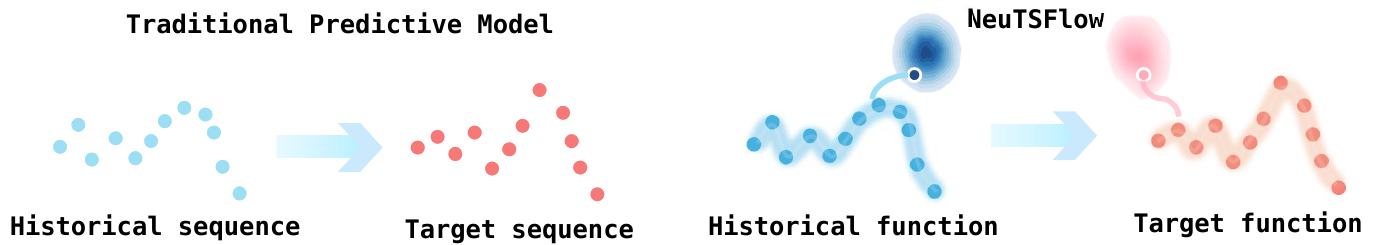} % 设置图片宽度为页面宽度
\caption{
Comparison between traditional time series forecasting and the NeuTSFlow paradigm. While traditional models learn mappings between discrete sequences, NeuTSFlow models the transition between the distributions of continuous functions corresponding to historical and future sequences, thereby capturing richer and more nuanced temporal dynamics.
%Comparison of Traditional Predictive Models and the NeuTSFlow Prediction Paradigm
}
\label{fig:enter-label}
\end{figure}
To tackle these challenges, \textit{we propose NeuTSFlow, a novel framework that utilizes Neural Operators to enable flow matching}. Neural Operators can endow the velocity field of flow matching with the ability to learn the relationships between functions. This in turn facilitates flow matching beyond merely learning the paths between probability distributions; rather, it can learn the measure paths between historical and future function families, introducing a function-family perspective to time series prediction tasks. 

Flow Matching is an advanced generative modeling~\cite{yuan2024diffusion,kong2020diffwave,lipman2023flowmatchinggenerativemodeling,tong2024improvinggeneralizingflowbasedgenerative} approach that enables connections between arbitrary distributions~\cite{albergo2023buildingnormalizingflowsstochastic}. By leveraging flow matching in function spaces \cite{kerrigan2023functionalflowmatching}, NeuTSFlow establishes a theoretical framework for learning the transition paths between historical and future function families. For time series forecasting, NeuTSFlow constructs a marginal measure path conditioned on initial and target functions, modeled as the limiting state of Gaussian paths, and reformulates the loss function and prediction target to align model outputs with forecasting objectives.

Neural Operators, originally developed for data-driven solutions to partial differential equations (PDEs), offer a significant advantage: they learn mappings from initial functions to solution functions without requiring explicit PDE priors \cite{li2021fourierneuraloperatorparametric,li2020neuraloperatorgraphkernel}. This flexibility makes them particularly suitable as velocity field models for flow matching, enabling the modeling of relationships between continuous functions underlying discrete time series data. By incorporating time series-specific features, such as non-stationarity and decomposable trend-seasonality, we design neural operators explicitly tailored for time series modeling. Furthermore, we establish that these neural operators serve as a continuous generalization of the classical time series model, DLinear, providing both robust theoretical underpinnings and strong practical relevance.

To validate the effectiveness of our model and the function-family perspective, we conduct extensive evaluations on eight widely-used datasets across three tasks: Conventional forecasting task, Time Series Super-Resolution, and Cross-Resolution Temporal Learning. Time Series Super-Resolution evaluates the model's ability to reconstruct high-resolution time series data from low-resolution inputs, while Cross-Resolution Temporal Learning assesses the model's capability to leverage high-resolution data to assist in forecasting. The results underscore the effectiveness of both our model and the proposed methodology.
Our  contributions are as follows.

1. We propose a novel framing of time series forecasting as learning transitions between historical and future function families, shifting from traditional point-wise dependency modeling to directly capturing function-level features and the fundamental characteristics of underlying processes.

2. We propose NeuTSFlow, a framework using neural operators as velocity field models for flow matching to connect historical and future function families. Through targeted design for time series, NeuTSFlow effectively captures relationships between continuous functions and discrete time series, enabling accurate forecasting.

3. We validate NeuTSFlow through experiments on eight forecasting datasets and three tasks evaluating different model capabilities, demonstrating its ability to capture continuous functions and deliver accurate predictions.

\section{Preliminaries}
To ground our proposed approach, we first revisit the foundational concepts behind time series forecasting. This includes a reframing of the prediction problem in function space, as well as key tools such as flow matching and neural operators.

\textbf{Reframing Time Series Forecasting and Relevant Notations.}
Let the time domain be a continuous interval \( \mathcal{T} = [a, b] \subset \mathbb{R} \), where multivariate observations are obtained at discrete time points \( \{t_i\}_{i=1}^n \subset \mathcal{T} \) with ordered timestamps \( t_1 < t_2 < \cdots < t_n \). Each observation at time \( t_i \) is a \( C \)-dimensional vector \( \boldsymbol{y}_{i} \in \mathbb{R}^C \). Define historical sequence \( \mathcal{X} = \{(t_{i_j}, \boldsymbol{y}_{i_j})\}_{j=1}^S \), where \( \{t_{i_j}\} \) is an ordered subsequence with \( a \leq t_{i_1} < \cdots < t_{i_s} \leq b \). Forecast sequence \( \mathcal{Y} = \{(t_{k_q}, \boldsymbol{y}_{k_q})\}_{q=1}^L \), where \( \{t_{k_q}\} \) satisfies \( t_{i_s} < t_{k_1} < \cdots < t_{k_l} \leq b \). Traditional methods formulate the prediction task as learning a mapping  
$
G: \mathbb{R}^{S \times C} \to \mathbb{R}^{L \times C}.
$
However, such predictive methods only learn the mapping between discrete sequences and fail to capture the continuous nature of the time series underlying the continuous processes. We now redefine time series prediction through the lens of continuous processes. Since finite discrete observations cannot uniquely determine a continuous function \( u \), and given that time series constitute noisy observations of the underlying continuous process, a function space framework is introduced to effectively model process uncertainty. Define a real separable Hilbert space \(\mathcal{F}\) of multivariate functions \(u\), where each function \(u\in\mathcal{F}\) is of the form:
$
u: \mathcal{T} \to \mathbb{R}^c, \quad t_k \mapsto u(t_k) = \big(u_1(t_k), \dots, u_c(t_k)\big)^\top.
$
To formalize the latent structure of continuous-time processes, we introduce a probability measure \( \mu \) on the measurable space \( (\mathcal{F}, \mathcal{B}(\mathcal{F})) \), where \( \mathcal{B}(\mathcal{F}) \) denotes the Borel \(\sigma\)-algebra induced by the norm topology of \( \mathcal{F} \). In the context of time-series prediction, we aim to model the transition between two probability measures: \(\mu_H\), which encapsulates the statistical properties of historical time-series functions, and \(\mu_{F}\), which represents the probability measure associated with predicted time-series functions. This transition captures the evolution of the underlying dynamics and uncertainty from the known historical data to the future, enabling us to make probabilistic predictions about the time - series values at the future time points.
$
M: \mathcal{\mu}_H \to \mathcal{\mu}_F.
$

% 由于连续时间域上的无限维函数空间无法由有限离散观测（我们的）唯一确定， 函数的先验结构（如光滑性、周期性）和内在随机性需要通过概率测度在函数空间中统一建模。噪声是这种变异性在观测端的体现之一

% Due to the presence of noise in the time - series data, we associate a probability measure \(\mu\) on the function space \(\mathcal{F}\), which describes the uncertainty and distribution of the underlying functions corresponding to the time - series observations.
% Let \(\mathcal{T}_{hist} = \{t_1, t_2, \ldots, t_{m}\} \subset \mathcal{T}\) (\(m < n\)) represent the historical sampling time points, and \(\mathcal{T}_{pred} = \{t_{m + 1}, t_{m + 2}, \ldots, t_{n}\}\) be the future time points for prediction. Denote \(\mathcal{F}_{hist}\) and \(\mathcal{F}_{pred}\) as the sub - spaces of \(\mathcal{F}\) restricted to \(\mathcal{T}_{hist}\) and \(\mathcal{T}_{pred}\) respectively, with their corresponding Borel \(\sigma\) - algebras \(\mathcal{B}(\mathcal{F}_{hist})\) and \(\mathcal{B}(\mathcal{F}_{pred})\).
\textbf{Flow Matching.}
% Flow matching (FM) is a simulation-free method for learning continuous normalizing flows that generates data by integrating an ordinary differential equation over a learned vector field. Here, we first give an overview of functional flow matching.
Let \(X_0 \sim p\) and \(X_1 \sim q\) be source and target distributions in \(\mathbb{R}^d\). Flow Matching (FM)~\cite{kerrigan2023functionalflowmatching} constructs a probability path \(p_t\) between \(p_0 = p\) and \(p_1 = q\) by learning a time-dependent velocity field \(u_t: \mathbb{R}^d \to \mathbb{R}^d\)~\cite{tong2024improvinggeneralizingflowbasedgenerative,lipman2023flowmatchinggenerativemodeling,albergo2023stochasticinterpolantsunifyingframework,albergo2023buildingnormalizingflowsstochastic,liu2022flowstraightfastlearning}. This field generates a flow \(\psi_t\) satisfying the ODE:
\(
\frac{d}{dt}\psi_t(x) = u_t(\psi_t(x)), \quad \psi_0(x) = x
\)
such that \(X_t = \psi_t(X_0) \sim p_t\). FM trains a neural network \(u_\theta\) via regression loss:\(
\mathcal{L}_{\text{FM}}(\theta) = \mathbb{E}_{t \sim \mathcal{U}[0,1], X_t \sim p_t} \|u_\theta(t,X_t) - u_t(X_t)\|^2
\).
% Inference involves solving the learned ODE from \(t=0\) to \(1\):
% \(
% X_1 = \text{ODESolve}(X_0, u_\theta, [0,1])
% \)
% where \(X_0\) is initialized from \(p\) and \(X_1\) constitutes generated samples. 
The flow matching framework on function spaces, as established in ~\cite{kerrigan2023functionalflowmatching,pooladian2023multisampleflowmatchingstraightening}, provides a comprehensive measure-theoretic foundation for flow matching. Defining a continuous flow \(\phi_t: [0, 1] \times \mathcal{F} \to \mathcal{F}\), governed by a time-dependent vector field \(v_t: [0, 1] \times \mathcal{F} \to \mathcal{F}\) via the ODE:
\(
\frac{\partial}{\partial t} \phi_t(g) = v_t(\phi_t(g)), \quad \phi_0(g) = g,
\)
where \(g \sim \mu_H\), the initial probability measure over \(\mathcal{F}\). The flow induces a probability path \(\mu_t = [\phi_t]_\# \mu_H\), the pushforward of \(\mu_H\) under \(\phi_t\). To further generalize, the marginal measure \(\mu_t\) is defined as:
\(\mu_t(A) = \int \mu_t^f(A) \, d\mu(f), \quad \forall A \in \mathcal{B}(\mathcal{F}),\)
where \(\mu(f)\) is a measure over functions and \(\mu_1^f\) is concentrated around \(f\). 

\textbf{Neural Operators.}
Neural operators\cite{li2020neuraloperatorgraphkernel,li2021fourierneuraloperatorparametric,Lu_2021} learn mappings between infinite-dimensional function spaces, enabling data-driven solutions for PDEs and functional regression. Let $\mathcal{F}$ be a function space (e.g., $L^2(\Omega)$) and $\mu$ a probability measure over $\mathcal{F}$. A neural operator $G_\theta: \mathcal{F} \to \mathcal{F}$ approximates an operator $G^\dagger$ by minimizing:  
\[
\mathcal{L}_{\text{NO}}(\theta) = \mathbb{E}_{f \sim \mu} \|G_\theta(f) - G^\dagger(f)\|_{\mathcal{F}}^2.  
\]  
Architectures like Fourier Neural Operators\cite{li2021fourierneuraloperatorparametric} parameterize $G_\theta$ efficiently in spectral domains.

\section{NeuTSFlow}
NeuTSFlow is a flow matching model for time series forecasting that leverages neural operators to capture the functional relationships between the family of functions associated with historical time series and those corresponding to target time series. The related work can be found in Appendix A.1, and the proofs for all propositions are provided in Appendix A.2.
\subsection{Framework Construction}
\begin{figure}
    \centering
    \includegraphics[width=0.8\linewidth]{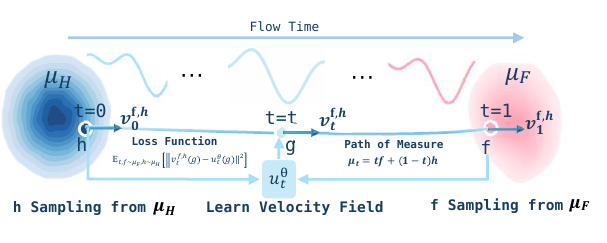}
    \caption{\textbf{Training overview of NeuTSFlow.} The model learns a velocity field \( u_t^\theta \) that aligns with the conditional velocity \( v_t^{f,h} \) between sampled functions \( h \sim \mu_H \) and \( f \sim \mu_F \) along the path of measure \( \mu_t = tf + (1 - t)h \).}
    \label{fig:enter-label}
\end{figure}
Given the probability measure \(\mu_H\) corresponding to the historical sequence and \(\mu_F\) corresponding to the predicted sequence, our objective is to construct a generative model capable of producing discrete samples from \(\mu_F\). To achieve this, we leverage flow matching to construct a continuous-time paths of probability measures \(\{\mu_t\}_{t \in [0,1]}\) that interpolates between \(\mu_H\) and \(\mu_F\). Specifically, this path is defined such that:  
\(
\mu_0 = \mu_H, 
\mu_1 = \mu_F.
\)

Given the probability measure \(\mu_H\), which corresponds to the initial historical sequence, where \(h \sim \mu_H\), \(h\) represents a time-series function sampled from \(\mu_H\). The flow-induced probability path satisfies \(\phi_t(g) \sim \mu_t\), with the relationship \(\mu_t = [\phi_t]_\# \mu_H\), where \([\phi_t]_\# \mu_H\) denotes the pushforward of \(\mu_H\) under the flow \(\phi_t\). Our goal is to learn such a flow that enables the construction of a continuous path connecting these probability measures.

\fcircle[fill=deblue]{2pt}  \textbf{Marginalization: Marginal Paths for Functional Flow Learning.}  
%While 
Direct estimation of the velocity field is intractable by design. However, the \textit{marginal path} can be constructed explicitly, enabling tractable approximation of the velocity field via functional integration.
%the marginal path can be explicitly constructed to derive its corresponding velocity field.
We therefore perform the following marginalization\cite{kerrigan2023functionalflowmatching}:

\begin{proposition}
Let $\mu$ be a probability measure on $\mathcal{F}$ and $\mu_t$ a family of measures on $G$ indexed by $t \in [0,1]$. Then the following hold: $\mu^{f,h}_t \ll \mu_t$ for $\mu \times \mu$-almost every $(f,h) \in \mathcal{F} \times \mathcal{F}$ and almost every $t \in [0,1]$. The integral
\(\int_0^1 \int_{\mathcal{F} \times \mathcal{F}} \int_G \|v^{f,h}_t(g)\| \, d\mu^{f,h}_t(g) \, d\mu(f)d\mu(h) \, dt < \infty\) is finite.  The function $v_t(g)$ is given by
\(v_t(g) = \int_{\mathcal{F} \times \mathcal{F}} v^{f,h}_t(g) \frac{d\mu^{f,h}_t}{d\mu_t}(g) \, d\mu(f) d\mu(h),\)
where $\frac{d\mu^{f,h}_t}{d\mu_t}$ denotes the Radon-Nikodym derivative of $\mu^{f,h}_t$ with respect to $\mu_t$.
\label{pro3.1}
\end{proposition}

Below, we focus on learning the marginal velocity field \(v^{f,h}_t\), where \(v^{f,h}_t : [0, 1] \times F \to F\). The key requirement for the marginal velocity field is its ability to map the function corresponding to the historical sequence to the function corresponding to the target sequence. To achieve this, we propose leveraging neural operators to effectively learn this mapping.

\fcircle[fill=deblue]{2pt}  \textbf{Measure-Path Construction for Infinite-Dimensional Flows. }
Given the well-established absolute continuity of Gaussian measures in infinite-dimensional (separable) Banach spaces \cite{nahmod4822absolute}, we leverage this property to construct a conditional path of measures in this work, defined as:
\begin{equation}
g_t = \mu_t^{f,h} = \mathcal{N}( tf + (1-t)h, \mathcal{C}_t), 
\label{phif,g}
\end{equation}
when \( \mathcal{C}_t \to 0 \), \( \mu_t^{f,h} \) approach to a bridge between the measures \( \mathcal{\mu}_H \) and \( \mathcal{\mu}_F \). This is a conditional path of measures that takes the endpoints as conditions. Since our initial distribution is not Gaussian, there exists a correspondence between the function representing a given historical discrete-time series and the function representing its target sequence during training on the dataset. Consequently, we construct a path of measures conditioned on both the initial and target functions, thereby providing the model with theoretical predictive capabilities~\cite{pooladian2023multisampleflowmatchingstraightening,liu2022flowstraightfastlearning,lipman2023flowmatchinggenerativemodeling}.  

Based on flow matching theory: \( \partial_t \phi_t(g) = v_t(\phi_t(g)) \), the associated velocity field is defined as:  
\begin{equation}
\lim_{\mathcal{C}_t \to 0} v_t^{f,h}(g) = f - h.
\label{velocityf,g}  
\end{equation}
This formulation results in a velocity field similar to the rectified flow proposed in \cite{liu2022flowstraightfastlearning}, where the optimal transport velocity field aligns with straight-line paths. Such paths leverage the shortest-path property to minimize the transport cost. Additionally, this approach offers a principled framework for characterizing measure transport in infinite-dimensional spaces while preserving the inherent properties of Gaussian distributions.

\fcircle[fill=deblue]{2pt} \textbf{Target Reparameterization for Functional Forecasting. }
Numerous studies~\cite{liu2022non,watson2023novo} in sequence prediction have investigated the reparameterization of prediction targets. Inspired by this, we design the output of the conditional velocity field to depend directly on the prediction target. As shown in~\eqref{phif,g}, predicting \(f\) is equivalent to predicting \(v_t^{f,h}(g)\), since \(h\) remains known during both training and inference. Thus, by defining \(v_t^{f,h}(g) = f\), the velocity field effectively becomes a mapping that predicts the target sequence as a function \(f\).

\subsection{Marginal Velocity Field Model $u_\theta^t$}
We now introduce a learnable model for the marginal velocity field $u_\theta^t$, which governs the transition between functions over time. This section details how the model is trained using discrete samples of functions and how it enables functional forecasting.

\fcircle[fill=deblue]{2pt} \textbf{Velocity Field Model Training and Inference. }
Based on~\eqref{phif,g} and \eqref{velocityf,g}, our velocity field is conditioned on the functions at both endpoints. Therefore, during training, the input consists of the historical sequence’s corresponding function \( h \), the function \( g \) along the path at time \( t \), and the time \( t \) itself. The target sequence’s corresponding function \( f \) serves as the training objective. Once this marginal velocity field is learned, inference becomes straightforward: given \( h \), the marginal velocity field at any time \( t \) can be computed. By solving the ODE \( \frac{\partial}{\partial t} \phi_t(g) = v_t(\phi_t(g)) \) using an ODE solver, the function \( g \) at time \( t \) can be determined. This process continues until \( t = 1 \), at which point \( f \) is obtained. The detailed training and inference procedure can be found in the Appendix A.3.

\fcircle[fill=deblue]{2pt} \textbf{Model Inputs. }
Our Marginal Velocity Field Model is built upon the theory of neural operators~\cite{li2021fourierneuraloperatorparametric,li2020neuraloperatorgraphkernel,Lu_2021}, which ensures that we can learn the underlying relationships between functions through discrete input values. Let the discretized input associated with \( g \) be \(\mathcal{G}_t \in \mathbb{R}^{L \times C}\), and the discretized history sequence associated with \( h \) be \(\mathcal{H} \in \mathbb{R}^{S \times C}\), which serves as a conditional input to the model. The discretized predicted time series corresponding to \( h \) is denoted as \(\mathcal{Y}\). 

To ensure that the velocity field operates as a function-to-function mapping, we design the model from the perspective of neural operators, effectively leveraging the inherent characteristics of time series data. First, to guaranteeensure dimensional consistency between the historical and predicted sequences during path construction, the historical embeddings are transformed throughusing a Multi-Layer Perceptron (MLP) to align witmatch the dimensionality of the predictedion data. This process is expressed as: \(\mathcal{H}_{\text{emb}} = \text{MLP}(\mathcal{H}) \in \mathbb{R}^{L \times C}\), where \(\mathcal{H} \in \mathbb{R}^{S \times C}\) represents the historical input, and \(\mathbf{H}_{\text{hist}}\) denotes the aligned historical embedding.

\fcircle[fill=deblue]{2pt} \textbf{Normalization}
To address the non-stationarity inherent in time series data, where the distribution evolves over time—a challenge not commonly faced in PDE-based solutions, we leverage the historical sequence to remove non-stationary components, specifically the mean and standard deviation. By incorporating these statistics as prior information for the prediction target, we enhance the model's performance and robustness in dynamic environments~\cite{kim2022reversible}.

To tackle this, we employ a normalization operator \(\mathcal{N}\) for the velocity field, defined as:
\(
\mathcal{N}(\mathcal{H}) = \frac{\mathcal{H} - \mu}{\sigma}, \quad \mu = \mathbb{E}[\mathcal{H}], \quad \sigma = \sqrt{\mathbb{V}[\mathcal{H}]},
\). The complete prediction process is then formulated as:
\begin{equation}
\mathcal{Y} = \mathcal{N}^{-1}[u_\theta^t(\mathcal{N}(\mathcal{H}),t,\mathcal{G}_t)].
\end{equation}

\fcircle[fill=deblue]{2pt} \textbf{Spectral Decomposition. } Time series data exhibit key characteristics such as long-term slow-varying trends, periodic patterns, and noise. To leverage these properties, we decompose the historical sequence \(\mathcal{H}\) in the frequency domain into trend and seasonal components.  
The proposed framework is implemented using the one-dimensional Discrete Fourier Transform (DFT), applied to each input \(\mathcal{H}\) as follows:  
\begin{equation}
Z = \text{DFT}(\mathcal{H}), \quad K = \text{TopK}(\text{Amp}(Z)), \quad \mathcal{H}_{\text{season}} = \text{IDFT}(\text{Filter}(K, Z)), \quad \mathcal{H}_{\text{trend}} = \mathcal{H} - \mathcal{H}_{\text{season}}.
\end{equation}
It illustrates the transformation of the input \(\mathcal{H}\) into its Fourier components, represented as \(Z \in \mathbb{C}^{S \times I}\), where \(\mathbb{C}\) denotes the set of complex values. The function \(\text{TopK}(\cdot)\) identifies the frequencies corresponding to the \(K\) largest amplitudes, calculated using the \(\text{Amp}(\cdot)\) function. The operation \(\text{Filter}(\cdot)\) isolates the selected frequencies \(K\) from \(Z\). The components \(\mathcal{H}_{\text{trend}}\) and \(\mathcal{H}_{\text{season}}\) are individually processed through their respective models, and the outputs are subsequently summed to produce the final result.                                                                                                                    

\fcircle[fill=deblue]{2pt} \textbf{Dimension Embedding. }  We concatenate the input state \(\mathcal{G}_t \in \mathbb{R}^{L \times C}\), history embedding \(\mathcal{H}_{\text{emb}} \in \mathbb{R}^{L \times C}\), and temporal encoding \(t \in \mathbb{R}\) along the channel dimension, constructing an augmented representation \(\mathbf{z}_0 = [\mathcal{G}_t; \mathcal{H}_{\text{emb}}; t] \in \mathbb{R}^{L \times (2C + 1)}\). However, this embedding may insufficiently capture the mapping between original infinite-dimensional function spaces \(\mathcal{F} \to \mathcal{G}\). We employ a dimension-expansion strategy through learnable linear projection:  First, we append a new dimension to \(\mathbf{z}_0\):  
\(
\mathbf{z}_{0,\text{expanded}} = \mathbf{z}_0 \oplus \mathbf{1}_{L \times (2C+1) \times 1} \in \mathbb{R}^{L \times (2C + 1) \times 1},\)
where \(\oplus\) denotes the addition of a singleton dimension. We then expand this dimension via a learnable projection matrix:  
\(
\mathbf{z}_1 = \mathbf{z}_{0,\text{expanded}} \cdot W_e \), where \( W_e \in \mathbb{R}^{1 \times k},\) yielding \(\mathbf{z}_1 \in \mathbb{R}^{L \times (2C + 1) \times k}\) with \(k \gg 1\).

\fcircle[fill=deblue]{2pt} \textbf{Spectral Temporal Learning. }

Our model realizes the neural operator framework through explicit correspondence with the integral form \( v(y) = \sigma\left(\int \kappa(y, x)a(x) \, \mathrm{d}x\right) \), where \(\kappa(y, x)\) is encoded by frequency-domain weights and \(a(x)\) by the input's spectral representation. This generalizes traditional MLP (finite summation) to continuous function spaces, forming the core of our continuous DLinear formulation.  

Given input signals \( \mathbf{z}_1 \in \mathbb{R}^{L \times (2C + 1) \times k} \), we first adjust the dimension order to align the temporal dimension \( L \) for processing, then apply Real-valued Fast Fourier Transform (RFFT):  
\(
\mathbf{X}_\mathcal{F} = \text{RFFT}(\mathbf{z}_1) \in \mathbb{C}^{(2C + 1) \times k \times M}, \)
where \( M = \min(m_{\text{max}}, L/2 + 1) \) and \( m_{\text{max}} \) is a predefined truncation length that controls the maximum number of retained frequency modes.  Learnable complex-valued weights \( \mathbf{W} \in \mathbb{C}^{k \times k \times M} \) act as the frequency-domain kernel \(\kappa(y, x)\), enabling cross-channel interaction via:  
\begin{equation}
\mathbf{Y}_\mathcal{F}[c,k,:m] = \sum_{i=1}^{k} \mathbf{X}_\mathcal{F}[c,i,:m] \odot \mathbf{W}[i,k,:m], 
\end{equation}
Here, element-wise multiplication (\(\odot\)) models \(\kappa(y, x) \cdot a(x)\) in frequency space, while summation over \(i\) emulates the integral aggregation across input features. Retaining low-frequency modes (\(m \leq M\)) performs spectral filtering, analogous to kernel smoothing in the integral operator. The inverse RFFT (IRFFT) transforms the spectrum back to the temporal domain, corresponding to the integral result \(\int \kappa(y, x)a(x) \, \mathrm{d}x\):  
\(\mathbf{Y_\mathcal{R}} = \text{IRFFT}(\mathbf{Y}_\mathcal{F}, n=L) \in \mathbb{R}^{L \times (2C + 1) \times k}.\)
Finally, a channel-wise linear projection maps the \((2C + 1) \times k\) features to \( C \) dimensions:  
\(\mathcal{Y} = \text{Linear}\left( \text{Flatten}(\mathbf{Y_\mathcal{R}}, \text{start\_dim}=1) \right) \in \mathbb{R}^{L \times C}\)
where the learnable projection weights aggregate information across channels and features to achieve the target dimensionality.

\fcircle[fill=deblue]{2pt}  \textbf{Function-to-Function Mapping Guarantee.}
Combining the aforementioned components, we arrive at the following proposition, which ensures the correctness of our neural operator theory and function space flow matching:
\begin{proposition}  
The marginal velocity field model \( u_\theta^t: \mathcal{F} \times [0,1] \to \mathcal{F} \) constitutes a neural operator that serves as the continuous-time version of DLinear~\cite{zeng2023transformers} and satisfies Function-to-Function Mapping. The operator \( u_\theta^t \) is decomposed into the following interconnected modules: normalization, spectral decomposition, dimension embedding, and spectral temporal learning.  
\end{proposition}

\fcircle[fill=deblue]{2pt} \textbf{Loss Function.}
To perform functional regression on the marginal vector field, we aim to approximate \( v_t(g) \) using a model \( u_t^\theta(g) \) with parameters \( \theta \in \mathbb{R}^p \). This can be achieved by minimizing the loss: \(L(\theta) = \mathbb{E}_{t \sim U[0,1], g \sim \mu_t} \left[ \|v_t(g) - u_t^\theta(g)\|^2 \right],\) where \( U[0,1] \) represents a uniform distribution over the interval \([0,1]\). The model \( u \) is defined as a mapping \( u : \mathbb{R}^p \times [0,1] \times F \to F \), i.e., a parameterized, time-dependent operator acting on the function space \( F \). However, directly minimizing \( L(\theta) \) is computationally intractable. Similar to \cite{lipman2023flowmatchinggenerativemodeling}, we consider the conditional loss:  
\begin{equation}
L_{\text{con}}(\theta) = \mathbb{E}_{t \sim U[0,1], f \sim \mu_F, h \sim \mu_H} \left[ \|v_t^{f,h}(g) - u_t^\theta(g)\|^2 \right],  
\end{equation} instead of regressing on the unknown \( v_t \), we regress on the known conditional vector fields \( v_t^{f,g} \). This proposition ensures the correctness of learning \( v_t^{f,h}(g) \).
\begin{proposition}  
Under the same conditions of Proposition \ref{pro3.1}, the gradients of the loss and the conditional loss are identical:  
\(
\nabla_\theta L(\theta) = \nabla_\theta L_{\text{con}}(\theta).
\)
\end{proposition}

\section{Experiment}
We empirically validate NeuTSFlow on multiple benchmark datasets and forecasting tasks. The goal is to assess its performance in modeling function-level transitions, as well as its ability to generalize across varying temporal resolutions and domains.

\textbf{Experiment Settings.}
To demonstrate the superiority of our model, we compared it against several representative baselines, including Transformer-based models such as iTransformer~\cite{liu2024itransformerinvertedtransformerseffective}, FedFormer~\cite{zhou2022fedformer}, Pathformer~\cite{chen2024pathformer}, and PatchTST~\cite{nie2022time}. Additionally, we evaluated our model against MLP-based approaches like DLinear~\cite{zeng2023transformers} and TimesMixer~\cite{wang2024timemixerdecomposablemultiscalemixing}, as well as other established models such as Autoformer~\cite{chen2021autoformer}, Pyraformer~\cite{liu2021pyraformer}, and Informer~\cite{zhou2021informer}. Due to space limitations and the need for prioritization, we selected the six most representative models for MSE comparison. Our experiments use eight real-world datasets, including ECL, four subsets of ETT, Exchange, Traffic, and Weather, which were previously employed in Autoformer \cite{chen2021autoformer}. The full results are provided in Appendix A.4. A detailed description of the relevant models and datasets can be found in Appendix A.5.
\subsection{Time Series Super-resolution}

\textbf{Task Description} The proposed model is validated on the time series super-resolution task, which is conceptually analogous to super-resolution in computer vision. In this context, resolution refers to the sampling frequency of time series data, with the objective of reconstructing high-sampling-rate sequences from low-sampling-rate observations \cite{gong2024superresolutionnetworktelemetrytime}.

\textbf{Task Setup}  To construct a realistic dataset pair with high-sampling-rate (HSR) and low-sampling-rate (LSR) characteristics for time series analysis, we employ established benchmark datasets—such as the ETTh1 dataset—as the HSR reference. The corresponding LSR dataset is generated through a fixed-interval decimation process, where data points are subsampled at regular temporal intervals. 
\begin{wrapfigure}{r}{0.6\textwidth} % 图片靠右（r），宽度占 0.5\textwidth
    \centering \vspace{-0.2cm}
    \includegraphics[width=0.6\textwidth]{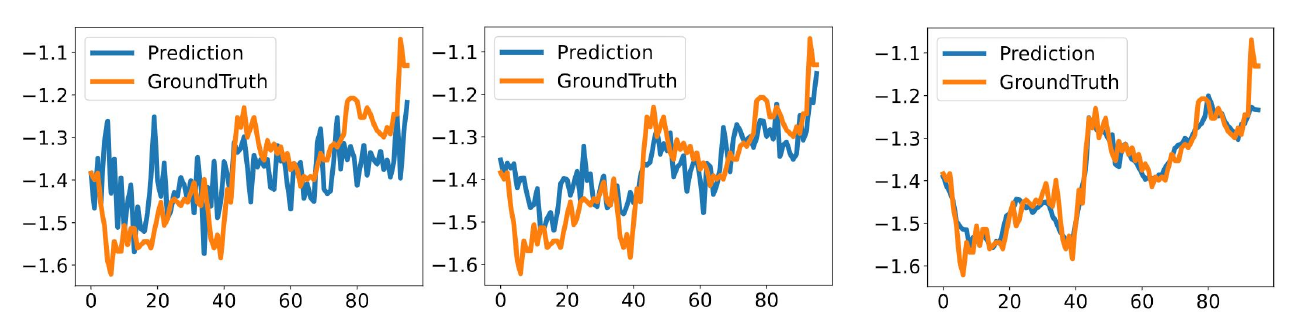} % 调整图片宽度
    \caption{From left to right: super-resolution results produced by DLinear, iTransformer, and NeuTSFlow.}
\end{wrapfigure}
Specifically, the LSR input sequence length is set to 24 time steps, derived from an HSR input sequence length of 96 time steps via a temporal decimation factor of 4. Both HSR and LSR sequences are strictly aligned to share identical temporal start and end, ensuring phase consistency and direct comparability across resolutions. This alignment is critical to preserve the temporal correspondence between datasets, avoiding misalignment errors that could compromise model training.  

\textbf{Task Results} as Table \ref{superresolution} NeuTSFlow consistently outperforms competing models, demonstrating superior time series super-resolution capabilities. It excels at recovering high-frequency components of the underlying continuous function from low-sampling-rate discrete points, addressing the core challenge of super-resolution tasks. This highlights NeuTSFlow's ability to model the continuous nature of time series, distinguishing it from traditional approaches like Dlinear. The results underscore the need to move beyond discrete point prediction in time series analysis and adopt continuous function modeling, which is vital for tasks such as high-resolution reconstruction and sensor interpolation.

\begin{table*}[t!]
\centering
\caption{Time series super-resolution performance (MSE) across eight benchmark datasets. Lower values indicate better performance. NeuTSFlow consistently achieves the \colorbox[HTML]{FFEAD9}{best results}, ranking first on all datasets and outperforming state-of-the-art baselines by a significant margin.}.
\def\arraystretch{1.01}
\resizebox{\textwidth}{!}{
\begin{tabular}{lcccccccc|c}
\toprule
\textbf{Model} & \textbf{Weather}   & \textbf{Traffic} & \textbf{ETTh1} & \textbf{ETTh2} & \textbf{ETTm1} & \textbf{ETTm2} & \textbf{Exchange} &\textbf{ECL} & \textbf{Avg Rank} \\
\midrule

\textbf{NeuTSFlow} &\cellcolor[HTML]{FFEAD9}\textbf{0.0231(1)}& \cellcolor[HTML]{FFEAD9}\textbf{0.2434(1)}   & \cellcolor[HTML]{FFEAD9}\textbf{0.0842(1)} &  \cellcolor[HTML]{FFEAD9}\textbf{0.0501(1)}  & \cellcolor[HTML]{FFEAD9}\textbf{0.0297(1)} & \cellcolor[HTML]{FFEAD9}\textbf{0.0211(1)}  & \cellcolor[HTML]{FFEAD9}\textbf{0.0024 (1)} & \cellcolor[HTML]{FFEAD9}\textbf{0.0791(1)}  &\cellcolor[HTML]{FFEAD9}\textbf{1.000(1)}\\

\textbf{iTransformer(2024)} &0.0256(4)& 0.2592(2)  &  0.0901(2) & 0.0537(3)  & 0.0319(2)  &  0.0228 (3)  & 0.0031(5) & 0.0804(3)& 3.000(2)\\

\textbf{FedFormer(2022)} &0.0417(7)& 0.3871(7)   &  0.1032(6)   & 0.0914(7)  & 0.0379(6) & 0.0242(7)  &   0.0049(7) & 0.0893(7) &6.750(7)\\
\textbf{DLinear(2023)} &0.0563(8)& 0.8295(8)   &  0.4136(8)& 0.3813(8)      & 0.1839(8)  &  0.0684(8)  & 0.0053(8)  & 0.3102(8) &8.000(8)\\

\textbf{TimeMixer(2024)}& 0.0268(6)& 0.2652(5)  &  0.1613(7)    & 0.0681(6) & 0.0389(7) & 0.0233(4)  &  0.0033(6) & 0.0868(6) &5.875(6)\\

\textbf{Pathformer(2024)}&0.0259(5)& 0.2613(4) &  0.0951(5) &  0.0533(2)  & 0.0321(3) & 0.0225 (2)  &   0.0028(2) & 0.0851(5) &3.500(4)\\

\textbf{PatchTST(2023)} &0.0254(3)& 0.2886(6)  &  0.0944(4)   & 0.0543(5) &0.0363(5) & 0.0234(5)  &  0.0028(2) & 0.0824(4)&4.250(5)\\
\textbf{Timexer(2025)} & 0.0241(2) &  0.2610(3)  &  0.0931(3)   & 0.0541(4) & 0.0323(4) & 0.0234 (5)  &  0.0029(4) & 0.0801(2) & 3.375(3)\\
\bottomrule
\label{superresolution}
\vspace{-0.3in}
\end{tabular}
}
\end{table*} 
%%%%
\begin{table*}[t!]
\centering
\caption{
%Cross-Resolution Temporal Learning
Evaluation of model performance on the Cross-Resolution Temporal Learning (CRTL) task. NeuTSFlow has the \colorbox[HTML]{FFEAD9}{best results} across all datasets, indicating its effectiveness in learning resolution-agnostic functional representations.
}
\def\arraystretch{0.9}
\resizebox{\textwidth}{!}{
\begin{tabular}{lcccccccc|c}
\toprule
\textbf{Model} & \textbf{Weather}   & \textbf{Traffic} & \textbf{ETTh1} & \textbf{ETTh2} & \textbf{ETTm1} & \textbf{ETTm2} & \textbf{Exchange} &\textbf{ECL} & \textbf{Avg Rank} \\
\midrule

\textbf{NeuTSFlow} & \cellcolor[HTML]{FFEAD9}\textbf{0.141(1)} & \cellcolor[HTML]{FFEAD9}\textbf{0.430(1)}   & \cellcolor[HTML]{FFEAD9}\textbf{0.370(1)} & \cellcolor[HTML]{FFEAD9}\textbf{0.316(1)}  & \cellcolor[HTML]{FFEAD9}\textbf{0.285(1)}       & \cellcolor[HTML]{FFEAD9}\textbf{0.172(1)}      & \cellcolor[HTML]{FFEAD9}\textbf{0.084(1)} & \cellcolor[HTML]{FFEAD9}\textbf{0.200(1)} & \cellcolor[HTML]{FFEAD9}\textbf{1.000(1)}\\

\textbf{iTransformer(2024)} &0.170(6) & 0.587(6)  & 0.457(6) & 0.349(5)  & 0.300(5) & 0.189(7)  & 0.131(6) & 0.243(6) &5.875(6)\\

\textbf{FedFormer(2022)} &0.220(8) & 0.595(7)  & 0.390(5) & 0.355(7)  & 0.385(8) & 0.205(8)  & 0.150(7) & 0.250(7) &7.125(7)\\

\textbf{DLinear(2023)}  &0.174(7) & 0.758(8)  & 0.485(8) & 0.443(8)  & 0.329(7) & 0.186(6)  & 0.177(8) & 0.306(8) &7.500(8)\\

\textbf{TimeMixer(2024)}&0.146(3) & 0.470(3)  & 0.375(2) & 0.334(3)  & 0.294(3) & 0.179(4)  & 0.093(3) & 0.211(2) &2.875(2)\\

\textbf{Pathformer(2024)}&0.165(5) & 0.485(5)  & 0.385(3) & 0.335(4)  & 0.295(4) & 0.173(2)  & 0.100(5) & 0.220(4) &4.000(4)\\

\textbf{PatchTST(2023)} &0.151(4) & 0.475(4)  & 0.460(7) & 0.350(6)  & 0.305(6) & 0.176(3)  & 0.087(2) & 0.239(5) &4.625(5)\\

\textbf{Timexer(2025)} &0.144(2) & 0.435(2)  & 0.386(4) & 0.329(2)  & 0.291(2) & 0.180(5)  & 0.094(4) & 0.212(3) &3.000(3)\\

\bottomrule
\label{CrossResolution}
\end{tabular}
}
\end{table*}

\subsection{Cross-Resolution Temporal Learning}
\textbf{Task Challenges and Design} We introduce a novel task named Cross-Resolution Temporal Learning (CRTL), designed to assess a model's capability of leveraging high-resolution time-series data to enhance prediction performance for low-resolution sequences. Temporal patterns exhibit distinct characteristics across different resolutions: for example, hourly-scale data captures daily periodicity, while daily-scale data reveals monthly trends. This resolution discrepancy leads to problems when applying models trained on high-resolution data to low-resolution prediction tasks.

To address this, we formalize the task as follows: during the training phase, low-resolution sequences serve as input, whereas high-resolution data are treated as the target. At inference time, the model generates low-resolution predictions by downsampling its high-resolution output predictions. Applications include leveraging high-frequency minute-level data to capture patterns of short-term stock price fluctuations, which can assist in predicting daily stock price trends. Moreover, high-frequency financial market data can enhance the accuracy of forecasting low-frequency macroeconomic indicators, offering valuable insights to support macroeconomic policy formulation \cite{galvao2022forecasting}.

\textbf{Task Setup} During training, the model takes a low-resolution time series of length 24 as input and predicts a high-resolution time series of length 96. During inference, the model predicts a 96-length high-resolution sequence from a 24-length low-resolution input, which is then downsampled by a factor of 4 to produce a 24-length output. Thus, the task maps a 24-length input to a 24-length output.

\textbf{Task Results} Our model achieves state-of-the-art results across all datasets, as shown in Table \ref{CrossResolution}. The core challenge of the CRTL task lies in leveraging high-resolution data to enhance the predictive capability of low-resolution sequences. NeuTSFlow addresses this by mapping low-resolution inputs to a high-resolution continuous space for feature refinement, followed by controlled downsampling to regress to the target resolution. Compared to traditional methods, the model demonstrates superior performance in cross-scale information transfer. Additional experiments and comparisons in Appendix A.6 further highlight the advantages of NeuTSFlow from the perspective of functional modeling.

\subsection{Conventional Forecasting Task}
\textbf{Task Setup} The input length is set to 96, while the output lengths are configured as 96, 192, 336, and 720. For clarity and ranking purposes, the main text only reports the MSE results for the output length of 96.

\textbf{Task Results} As shown in Table \ref{forecasting}, NeuTSFlow outperforms in traditional time series forecasting tasks, especially on datasets with strong periodicity and structured patterns like ETT, Exchange, and ECL. While it does not always rank first, it remains highly competitive across all datasets, showcasing its robustness and versatility.

\begin{table*}[t]
\centering
\caption{
Mean Squared Error (MSE) results for the conventional time series forecasting task across eight benchmark datasets. NeuTSFlow consistently ranks among the \colorbox[HTML]{FFEAD9}{top-performing models}, achieving the lowest average rank and outperforming established Transformer-based and MLP-based baselines on most datasets.
%Conventional Forecasting Task
}.
\def\arraystretch{0.9}
\resizebox{\textwidth}{!}{
\begin{tabular}{lcccccccc|c}
\toprule
\textbf{Model} & \textbf{Weather}   & \textbf{Traffic} & \textbf{ETTh1} & \textbf{ETTh2} & \textbf{ETTm1} & \textbf{ETTm2} & \textbf{Exchange} &\textbf{ECL} & \textbf{Avg Rank} \\
\midrule

\textbf{NeuTSFlow} & 0.160(2) & 0.428(2)   & \cellcolor[HTML]{FFEAD9}\textbf{0.367(1)} &  \cellcolor[HTML]{FFEAD9}\textbf{0.280(1)}  & \cellcolor[HTML]{FFEAD9}\textbf{0.314(1)} & \cellcolor[HTML]{FFEAD9}\textbf{0.170(1)}  & \cellcolor[HTML]{FFEAD9}\textbf{0.081(1)} & \cellcolor[HTML]{FFEAD9}\textbf{0.138(1)} & \cellcolor[HTML]{FFEAD9}\textbf{1.250(1)}\\

\textbf{iTransformer(2024)} & 0.178(6) & \cellcolor[HTML]{FFEAD9}\textbf{0.426(1)}  &  0.386(6) & 0.301(5)  & 0.334(5)  &   0.181(5) &   0.086(4)     & 0.148(5) &4.625(5)\\

\textbf{FedFormer(2022)} & 0.215(8) & 0.588(7)   &  0.376(3)   & 0.347(8)  & 0.379(8) & 0.205(8)  & 0.148(7)  & 0.181(6) &6.875(8)\\
\textbf{DLinear(2023)} & 0.195(7) &  0.613(8) &  0.386(6) & 0.335(7)      & 0.345(7)  &  0.193(7)  & 0.088(5)  & 
0.185(7)  &6.750(7)\\

\textbf{TimeMixer(2024)}& 0.165(4) & 0.461(4) &   0.374(2)  & 0.294(4)  & 0.331(4) & 0.175(4)  &  0.083(2) & 0.143(3) & 3.375(3)\\

\textbf{Pathformer(2024)}& 0.163(3)   & 0.479(6) &  0.382(5) &  0.283(2)  & 0.319(3) & 0.172(2)  &   0.085(3) & 0.146(4)  &3.500(4)\\

\textbf{PatchTST(2023)} & 0.173(5) &  0.466(5)  &  0.383(8)   & 0.323(6) &0.336(6) & 0.184(6)  &  0.094(6) & 0.191(8) & 6.250(6)\\

\textbf{Timexer(2025)} & \cellcolor[HTML]{FFEAD9}\textbf{0.157(1)} &  0.430(3)  &  0.381(4)   & 0.286(3) &0.318(2) & 0.173(3)  &  0.088(5) & 0.140(2) & 2.875(2)\\

\bottomrule
\label{forecasting}
\vspace{-0.3in}
\end{tabular}
}
\end{table*}

%\vspace{-2ex}

\subsection{Ablation studies}
To validate the contribution of key modules in NeuTSFlow, we conduct ablation experiments by removing four critical components: flow matching, neural operator, normalization layer, and spectral decomposition layer. The experiments are performed on three representative datasets across three tasks. The results are reported in Table \ref{ablation} as MSE values. "W/O neural operator" indicates replacing the Marginal Velocity Field Model with the DLinear model. "W/O flow matching" refers to using only the Marginal Velocity Field Model for prediction. As shown in Table \ref{ablation}, the removal of the Neural Operator results in the largest performance drop, underscoring its critical role in modeling continuous temporal dynamics and the importance of a function-based perspective. Similarly, the removal of Flow Matching leads to the second-largest impact, highlighting its effectiveness in capturing transitions between function families. The removal of the Normalization Layer primarily affects non-stationary data, demonstrating its significance in managing dynamic variability.  Overall, the Neural Operator and Flow Matching are the core innovations behind NeuTSFlow’s superior performance. These results validate the utility of each component and support the  rationale for integrating discrete-domain operations with continuous-time modeling, yielding to a robust framework for time series forecasting.

\begin{table*}[t]  
\centering  
\caption{ Ablation results across three tasks. Neural operators and flow matching are most critical, while normalization and spectral decomposition also boost performance. CFT=Conventional Forecasting Task, CRTL= Cross-Resolution Temporal Learning, and TSSR=Time Series Super-Resolution }  
\def\arraystretch{0.9}  
\resizebox{\textwidth}{!}{  
\begin{tabular}{lcccccc}  
\toprule  
\multirow{2}{*}{\textbf{Task}} & \multirow{2}{*}{\textbf{Dataset}} & \multicolumn{5}{c}{\textbf{Model Variant}} \\  
\cmidrule{3-7}  
& & \textbf{NeuTSFlow} & \textbf{w/o neural operator} & \textbf{w/o flow matching} & \textbf{w/o normalization} & \textbf{w/o spectral decomposition} \\  
\midrule  
\multirow{3}{*}{CFT}  
& ETTh1 & 0.367 & 0.384 & 0.379 & 0.376 & 0.372 \\  
& Weather  & 0.160 & 0.188 & 0.179 & 0.172 & 0.168 \\  
& Exchange  & 0.081 & 0.086 & 0.085 & 0.083 & 0.084 \\  
\midrule  
\multirow{3}{*}{CRTL }  
& ETTh1 & 0.370 & 0.442 & 0.427 & 0.405 & 0.418 \\  
& Weather & 0.141 & 0.167 & 0.165 & 0.158 & 0.153 \\  
& Exchange & 0.084 & 0.117 & 0.099 & 0.092 & 0.105 \\  
\midrule  
\multirow{3}{*}{TSSR}  
& ETTh1 & 0.0842 & 0.1182 & 0.1021 & 0.0997 & 0.0910 \\  
& Weather & 0.0231 & 0.0386 & 0.0303 & 0.0272 & 0.0254 \\  
& Exchange & 0.0024 & 0.0041 & 0.0030 & 0.0032 & 0.0028 \\  
\bottomrule  
\label{ablation}  
\end{tabular}  
}  
\end{table*}
\vspace{-2ex}

\section{Conclusion}
This work addresses the limitation of current time series forecasting models that treat data as discrete sequences, neglecting their origin as noisy samples of continuous processes. We propose a novel perspective that redefines time series forecasting as the task of learning transitions between historical and future function families. This paradigm shifts the focus from conventional point-wise dependency modeling to directly capturing function-level representations and the fundamental dynamics of the underlying processes. Building on this framing, we propose NeuTSFlow, a framework that leverages neural operators as velocity field models for flow matching to connect historical and future function families. By designing marginal measure paths, reformulating loss functions, and adapting to the features of time series, NeuTSFlow effectively captures the relationships between continuous functions and discrete time series. We validate the effectiveness of NeuTSFlow through experiments on eight diverse forecasting datasets and three tasks assessing model capabilities. The results demonstrate NeuTSFlow's ability to reliably capture continuous functions and deliver accurate predictions, offering a promising direction for advancing time series forecasting methodologies.

\section{Acknowledgement}
CWC is supported by the Swiss National Science Foundation under grant number 20HW-1 220785. AIAR gratefully acknowledges the support of the Yau Mathematical Sciences Center, Tsinghua University. This work is also supported by the Tsinghua University Dushi Program.

\bibliographystyle{abbrvnat}
\bibliography{cite}

\end{document}